\documentclass[10pt,twocolumn,letterpaper]{article}

\usepackage{iccv}
\usepackage{times}
\usepackage{epsfig}
\usepackage{graphicx}
\usepackage{amsmath}
\usepackage{amssymb}
\usepackage{multirow}
\usepackage{diagbox}
\usepackage{array}
\usepackage{booktabs}
\usepackage{enumitem}
\usepackage[abs]{overpic}
\usepackage{float}
\renewcommand{\vec}[1]{\boldsymbol{#1}}
\newcommand{\mat}[1]{\mathbf{#1}}

\newcommand{\pose}[0]{\vec{\theta}}
\newcommand{\shape}[0]{\vec{\beta}}

\newcommand{\loss}[0]{\mathcal{L}}
\usepackage[pagebackref=true,breaklinks=true,letterpaper=true,colorlinks,bookmarks=false]{hyperref}

\iccvfinalcopy %

\def\iccvPaperID{6488} %

\ificcvfinal\fi

\newcommand{\frl}{ClothSeq}

\begin{document}

\title{Neural-GIF: Neural Generalized Implicit Functions \\for Animating People in Clothing}

\author{Garvita Tiwari \textsuperscript{1,2} \qquad Nikolaos Sarafianos \textsuperscript{3}\qquad Tony Tung \textsuperscript{3} \qquad Gerard Pons-Moll\textsuperscript{1,2}\\\\
{\small \textsuperscript{1}University of Tübingen,  Germany, \qquad \textsuperscript{2}Max Planck Institute for Informatics, Saarland Informatics Campus, Germany}  \\
{\small\textsuperscript{3}Facebook Reality Labs, Sausalito, USA}\\
{\tt\scriptsize gtiwari@mpi-inf.mpg.de,  \{nsarafianos, tony.tung\}@fb.com, gerard.pons-moll@uni-tuebingen.de}}

\makeatletter
\let\@oldmaketitle\@maketitle%
\renewcommand{\@maketitle}{
	\@oldmaketitle%
	\begin{center}
	   	\begin{overpic}[width=0.48\textwidth,unit=1mm]{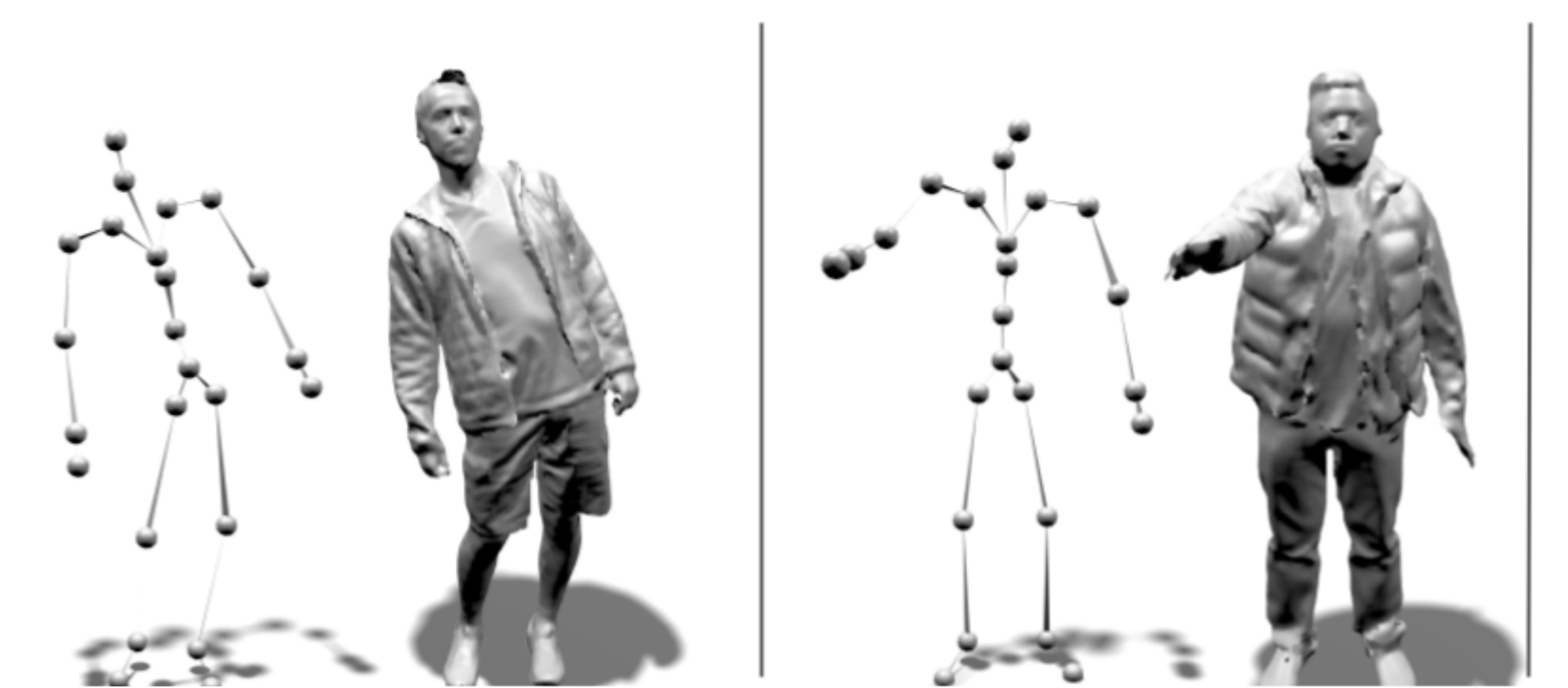}
   	\put(0,40){{\parbox{0.4\linewidth}{%
    Input pose }}}
    	\put(20,40){\colorbox{white}{\parbox{0.4\linewidth}{%
    Output  }}}
   	\put(45,40){{\parbox{0.4\linewidth}{%
    Input pose }}}
    	\put(65,40){\colorbox{white}{\parbox{0.4\linewidth}{%
    Output  }}}
\end{overpic}
	   	\begin{overpic}[width=0.48\textwidth,unit=1mm]{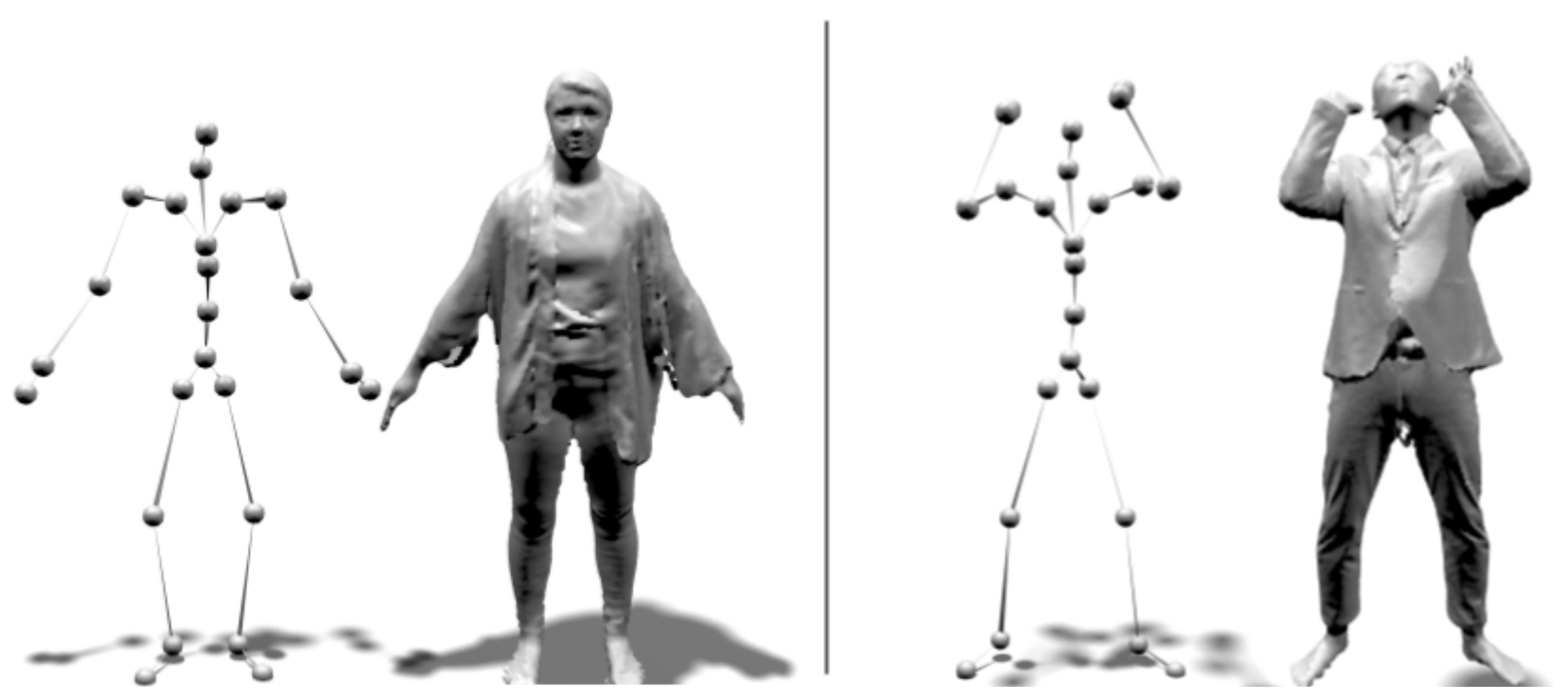}
   	\put(5,40){{\parbox{0.4\linewidth}{%
    Input pose }}}
    	\put(25,40){\colorbox{white}{\parbox{0.4\linewidth}{%
    Output  }}}
   	\put(50,40){{\parbox{0.4\linewidth}{%
    Input pose }}}
    	\put(70,40){\colorbox{white}{\parbox{0.4\linewidth}{%
    Output  }}}
\end{overpic}

	\end{center}
    \refstepcounter{figure}\normalfont Figure~\thefigure. We present \emph{Neural Generalized Implicit Functions} (Neural-GIF), to animate people in clothing as a function of body pose. Neural-GIF learns directly from scans, models complex clothing and produces pose-dependent details for realistic animation. We show for four different characters the query input pose on the left (illustrated with a skeleton) and our output animation on the right.
	\label{fig:teaser}
	\newline
}
\makeatother

\maketitle

\ificcvfinal\thispagestyle{empty}\fi

\begin{abstract}
    We present \emph{Neural Generalized Implicit Functions} (Neural-GIF), to animate people in clothing as a function of the body pose. Given a sequence of scans of a subject in various poses, we learn to animate the character for new poses. Existing methods have relied on template-based representations of the human body (or clothing). However such models usually have fixed and limited resolutions, require difficult data pre-processing steps and cannot be used with complex clothing. We draw inspiration  from  template-based  methods,  which  factorize motion into articulation and non-rigid deformation, but generalize this concept for implicit shape learning to obtain a more flexible model. We learn to map every point in the space to a canonical space, where a learned deformation field is applied to model non-rigid effects, before evaluating the signed distance field. Our formulation allows the learning of complex and non-rigid deformations of clothing and soft tissue, without computing a template registration as it is common with current approaches. Neural-GIF can be trained on raw 3D scans and reconstructs detailed complex surface geometry and deformations. Moreover, the model can generalize to new poses. We evaluate our method on a variety of characters from different public datasets in diverse clothing styles and show significant improvements over baseline methods, quantitatively and qualitatively. We also extend our model to multiple shape setting. To stimulate further research, we will make the model, code and data publicly available at~\cite{project}.

\end{abstract}

\section{Introduction}
Human avatars enable numerous applications related to augmented and virtual reality, such as telepresence for enhanced communication and entertainment, and have been instrumental to reconstruct and perceive people in images~\cite{alldieck19cvpr,alldieck2018video,bogo2016keep,habermann20deepcap,xu2019denserac,xiang2019monocular}. Human shape deforms according to articulation, soft-tissue and non-rigid clothing dynamics, which make realistic animations extremely challenging.

State-of-the-art body models~\cite{SMPL:2015,SMPL-X:2019,xiang2019monocular} typically learn to deform a \emph{fixed topology template}, usually using linear blend skinning to model articulation, and blendshapes to model non-rigid effects~\cite{SMPL:2015}, even including soft-tissue~\cite{pons2015dyna} and clothing~\cite{ma20autoenclother,patel2020,santesteban2019virtualtryon}.
The use of a fixed template limits the type of clothing and dynamics that can be modeled. For example, it would be difficult to model the subjects in Fig.~\ref{fig:teaser} with one or more predefined templates. Furthermore, every type of deformation (soft-tissue or clothing) requires a different model formulation. Additionally for training a model, 3D/4D scans need to be brought into correspondence~\cite{SMPL:2015,pons2015dyna}, which is a challenging task especially for clothing~\cite{ponsmoll2017clothcap}, and not even well defined when garments vary in topology~\cite{bhatnagar2019mgn}.
Recent works have leveraged implicit function representation to reconstruct human shape from images~\cite{saito2019pifu,PifuHD} or 3D point-clouds~\cite{bhatnagar2020ipnet, chibane20ifnet}. These reconstructions are however static and not animatable. 

In this work, we propose a new model, called Neural Generalized Implicit Functions (Neural-GIF) to animate people in clothing as a function of body pose.
We demonstrate that we can model complex clothing and body deformations at a quality not easily achieved with a traditional fixed template-based mesh representations. %
In contrast to most prior work, we learn pose-dependent deformations without the need of registration of any pre-defined template, which degrades the resolution of the observations, and is a notoriously complex step prone to inaccuracies for complex clothing~\cite{ponsmoll2017clothcap}. Instead we solely require the pose of the input scans as well as the SMPL shape parameter (\(\beta\)). 
Another key advantage of our method is that it can represent different topologies using the exact same formulation --we show how to animate jackets, coats, skirts and soft-tissue of undressed humans. 
Neural-GIF consists of a neural network to approximate the signed distance field (SDF) of the posed surface. Naively learning to predict SDF from pose is hard. 
Instead, we draw inspiration from template-based methods which factorize motion into articulation and non-rigid deformation, but generalize this concept for implicit shape learning. 
Specifically, we learn to map every point surrounding the surface to a canonical space, where a learned deformation field is applied to model non-rigid effects, before evaluating the SDF. Our model (and its name) is inspired by the seminal paper~\cite{generalised_implicit}, which shows that a wide variety of shapes can be obtained by simply applying deformation fields to a base implicit shape. The advantage of Neural-GIF is that the network can more easily learn a base shape in canonical space, which can be deformed.   
In summary, our contributions are:
\begin{itemize}[leftmargin=*]
	\item [$\bullet$] Neural-GIF, an implicit based re-posable character, which can be directly learned from 3D scans. Our model can represent complex character/clothing scans of varied topology and geometry. 
		\item [$\bullet$] We introduce a canonical mapping network, which learns continuous skinning field in 3D space and unpose 3D points surrounding the scan to a canonical T-pose, without explicit supervision. 
	\item [$\bullet$] We introduce a \emph{displacement} field network, which shifts points in the canonical space before evaluating the SDF, yielding in fine details and deformation.
\end{itemize}

We test our method on a variety of scans originating from different datasets and provide extensive quantitative and qualitative comparisons. %
We also extend our formulation to multiple shape setting, by adding a shape dependent displacement field network.
Experiments demonstrate that our method generalizes to new poses, model complex clothing, and is significantly more robust and more detailed than existing methods~\cite{Saito:CVPR:2021, LEAP:CVPR:21, deng2019neural}.

\section{Related Work}
\label{sec:related_work}
\begin{figure*}[t]
    \centering
    \includegraphics[width=0.98\textwidth]{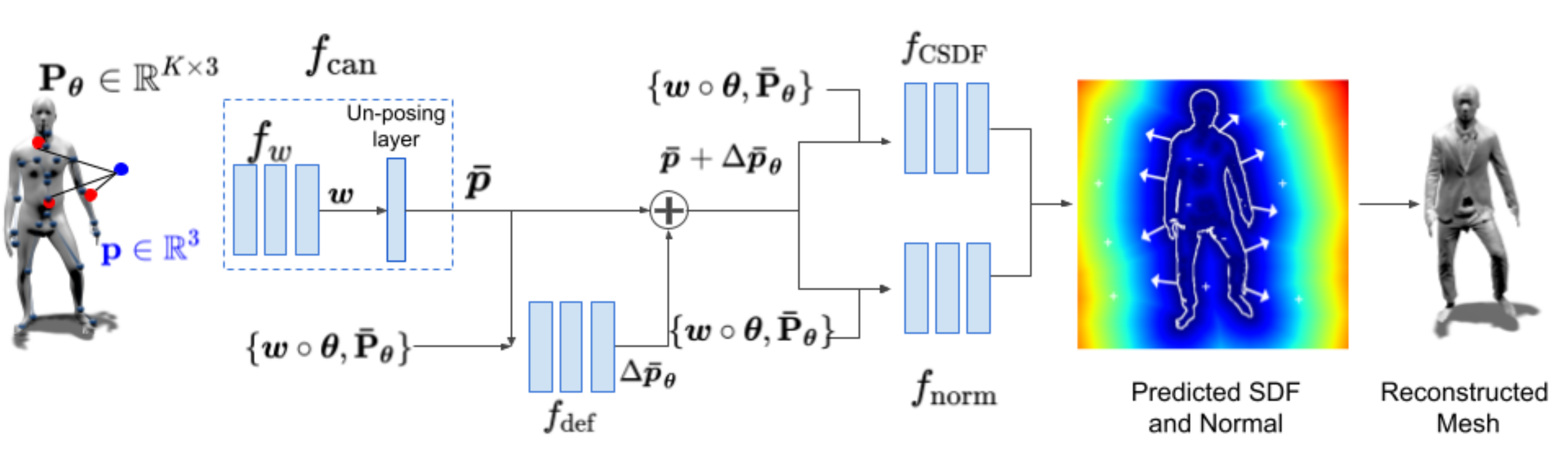}  \caption{We propose \emph{Neural Generalized Implicit Function} (Neural-GIF), which first maps every point $\vec{p}$ in 3D posed space to the canonical space by learning the associations between the point and the human body with $f_{\vec{w}}$ based on a pose encoding ($\mathbf{P}_{\pose}$) as input. Then we predict a displacement field model ($f_{\mathrm{def}}$) learns a pose-dependent deformation field in canonical space ($\Delta \vec{\bar{p}_{\mathrm{\pose}}}$), conditioned on weighted pose parameters ($w \circ \pose$) and pose encoding ($\mathbf{\bar{P}}_{\pose}$) in the canonical space. We add the displacement to the unposed point before evaluating the SDF using our canonical SDF network ($f_{\mathrm{CSDF}}$). We also employ a normal prediction network ($f_{\mathrm{norm}}$) to obtain results with fine-level details. Both $f_{\mathrm{CSDF}}$ and $f_{\mathrm{norm}}$ take weighted pose and pose encoding as input.}
    \label{fig:method}
\end{figure*}
\noindent\textbf{Human and Clothing models}: 
The problem of animating clothing/characters has been tackled using Physics-Based Simulation (PBS) methods for a long time. %
3D meshes are animated using PBS governed by some fundamental forces and collision detection. PBS methods require designing a garment/mesh template and manual intervention for finding suitable physical parameters. Past work has explored ways to automate this limitation by trying to infer these physical parameters mechanically from data~\cite{Miguel2012clothsim,Wang:2011:DDE}, or use videos to infer them~\cite{RosenhahnMVA2006,stoll_videobased}. PBS methods are also computationally expensive and the complexity increases for realistic animations. Several approaches generate realistic and high-frequency details by computing a high resolution mesh from a base coarse mesh using constraint-based optimization~\cite{SCA:SCA10:085-091,rohmer:hal-00516411}. 
One way of animating characters is by learning a transformation between pose-shape and the body surface and one can generate characters in new pose and body shape using this linear transformation~\cite{10.1145/1073204.1073207,SMPL:2015,STAR:2020}. Since such body models do not model personalized and clothing details, various methods~\cite{alldieck2019tex2shape,bhatnagar2019mgn,ponsmoll2017clothcap,jiang2020bcnet} have explored ways to extend the underlying body model~\cite{SMPL:2015}, by predicting per-vertex displacements for personalized and clothing features. However, they do not take pose-dependent deformations into account. Dyna~\cite{pons2015dyna}, extends the body model to predict soft-tissue dynamics caused by motion by learning a second-order auto-regressive model.
With the advances in deep learning and 3D data acquisition~\cite{tung08cvpr,tung2009iccv,tung12}, data-driven models of clothing and humans~\cite{xu2019denserac} are becoming increasingly popular. One key aspect in modeling human and clothing is articulation and pose-dependent non-rigid deformations. A significant amount of prior work models these deformations conditioned on the body shape-pose~\cite{bertiche2019cloth3d,DRAPE2012,santesteban2019virtualtryon}, the body pose~\cite{laehner2018deepwrinkles}, the body shape-clothing size/style~\cite{tiwari20sizer,garmentdesign_Wang_SA18} and the body shape-pose-clothing style~\cite{patel2020}. All these methods use a template-specific model, \ie either they are learned from offline PBS data~\cite{bertiche2019cloth3d,DRAPE2012,patel2020,santesteban2019virtualtryon,garmentdesign_Wang_SA18, gundogdu19garnet,vidaurre2020virtualtryon} or learned from template-specific registration of scans~\cite{laehner2018deepwrinkles,ma20autoenclother,tiwari20sizer}. ~\cite{zhang2020deep} generates fine details for any clothing styles, but requires low resolution meshes as input. This strictly limits the usability of such approaches to new clothing styles, especially to those with complex geometries and does not follow the body topology. There is no existing work that can model complex clothing like jackets, directly from scans.

\noindent \textbf{Modeling using Implicit representations}:
There is a plethora of recent methods~\cite{chibane20ifnet,chibane2020ndf,occ_net,Park_2019_CVPR} that explore neural implicit functions to represent 3D objects, including detailed articulated objects like human bodies and garments~\cite{chibane20ifnet,chibane2020ndf} but do not model pose-conditioned occupancy. Corona \etal~\cite{coronaSMPLicit} introduces a generative model for garments with different topology, but doesn't model fine pose dependent deformations. Like us, NASA~\cite{deng2019neural} extends the implicit surface formulation to represent articulated objects using a composition of part-based neural implicit functions. However the part-based representation in~\cite{deng2019neural} yields artifacts at the interfaces and does not capture fine details. Similar to our work, concurrent works like LEAP~\cite{LEAP:CVPR:21}, NiLBS~\cite{jeruzalski2020nilbs} and SCANimate~\cite{Saito:CVPR:2021}, propose a neural implicit model which extends the idea of skinning weights to volumetric skinning weights.  Previous work like ARCH~\cite{huang2020arch} and LoopReg~\cite{bhatnagar2020loopreg} also extend skinning weights to 3D space, but they are not learned and instead are pre-determined based on nearest neighbour. Prior work like~\cite{10.1145/3306346.3322969, bertiche2021deepsd} learn skinning weights for different topolgies using graph convolutions. Yang \etal~\cite{yang2021s3}, propose to predict shape, skeleton and skinning jointly from image or depth data to animate characters.
Also, in~\cite{huang2020arch}, clothed humans are modeled as pixel-aligned implicit functions in the unposed canonical space but do not model pose-dependent effects. Božič \etal~\cite{bo2020neural} model non rigid deformations using an implicit deformable shape via an identity specific deformation graph but their focus is tracking and reconstruction of non-rigid 3D objects. In~\cite{yenamandra2020i3dmm}, an implicit morphable model for faces is proposed, where shape is decoupled into an implicit reference shape and a deformation of this reference shape.

Another line of work in implicit scene representation comes from NeRF~\cite{mildenhall2020nerf}, where the focus is photorealistic rendering of a scene. Dynamic NeRF methods~\cite{park2020nerfies,pumarola2020dnerf} add the dynamic aspect to neural radiance fields, by first encoding the scene into canonical space, using a deformation field and then predicting density and view-dependent color values. Hence they can reconstruct and render novel images of objects under rigid and non-rigid motions from a single camera moving around the scene. In contrast, we learn a model which generalizes to new poses, learns from 3D scans, and integrates the SMPL kinematic skeleton.

In the seminal work of~\cite{generalised_implicit}, a generalized formulation of implicit functions is introduced to obtain new shapes by deforming and transforming an implicit base shape, by deforming and displacing points. Instead of using predefined deformation matrices and displacement maps, we impose articulated structure from the SMPL body model~\cite{SMPL:2015}, learn to map points to a canonical space, and learn a pose-dependent displacement field with neural networks.

\section{Method}

Our method, called Neural-GIF, is a neural model based on generalized implicit functions~\cite{generalised_implicit} for animating people in clothing. 
Given a pose($\pose$) as input, Neural-GIF predicts SDF, whose zero level set represents the character's surface $\mathcal{S}$. 
Without loss of generality, we use SMPL~\cite{SMPL:2015} for our pose representation. To train Neural-GIF we require sequences of 3D scans of a subject in a fixed clothing, and the corresponding SMPL parameters. We do not require non-rigid registration and un-posing as a pre-processing step~\cite{patel2020,ponsmoll2017clothcap,pons2015dyna,laehner2018deepwrinkles}, which is a tedious task and prone to error. 
Since learning the mapping $\pose \mapsto \mathrm{SDF}(\mathcal{S})$ is hard (large variations due to articulation, fine non-rigid deformations due to soft-tissue and clothing) and often results in missing limbs, we factorize the motion to learn a \emph{deformable SDF in a canonical} (unposed) space of the character. Three neural functions are composed to obtain the pose dependent SDF: 
\begin{itemize}[leftmargin=*]
    \item \textbf{Canonical mapping network}~(Sec.~\ref{canonical_net}): maps every point in the posed 3D space to the canonical unposed space by learning point to human body associations. 
    \item \textbf{Displacement field network}~(Sec.~\ref{def_net}):
     models non-rigid pose-dependent deformations (soft-tissue, cloth dynamics). Specifically, it predicts a continuous displacement field for points in the canonical space.%
    \item \textbf{Canonical SDF}~(Sec.~\ref{sdf_net}):~By composition, \emph{Canonical SDF} network, takes the transformed point with the above networks, and a pose encoding as input to predict the desired \emph{signed distance} for every query point. We additionally predict a surface normal field as a function of pose in canonical space to add realism in the results.
\end{itemize}

\subsection{Canonical Mapping Network}
\label{canonical_net}
Our canonical space is characterized by a T-pose of character. Learning in this space allows the network to discover a mean shape that can be deformed and articulated, which helps in pose generalization. Formally, the main task of the network is to map every point in the posed space (including off-surface points) to the canonical space $\vec{p}\mapsto \bar{\vec{p}}$. This requires associating the point $\vec{p}$ with the body, and inverting the articulated motion. 
While we could associate every point to its closest SMPL body model point, this is often ambiguous in occluded regions near the armpits or between the legs(See fig.~\ref{fig:our_variant}). Consequently, we propose to learn this \emph{association as a function of pose}.  

Instead of simply using pose $\pose$ as input, we use a body-aware pose representation. First, we take $K$ joints the SMPL mesh ($\vec{j_i}, i \in \{1 \hdots K \}$), and then define a pose encoding $\mathbf{P}_{\pose} \in \mathbb{R}^{K \times 3}$, where $\mathbf{P}_{\pose, i} = ||\vec{p} - \vec{j_i}||$. 

Given a query point in space and its pose encoding, the \emph{canonical mapping network} predicts the blend weights $\vec{w} \in \mathbb{R}^{K}$ of query point $\vec{p}$ and transforms it to the canonical space using a differentiable un-posing layer: %
\setlength{\abovedisplayskip}{3pt}
\setlength{\belowdisplayskip}{3pt}
\begin{eqnarray}
\label{eq:can_trans}
\vec{w} &=& f_{\vec{w}}(\vec{p}, \mathbf{P}_{\pose})\\
\label{eq:can_trans2}
 \vec{\bar{p}} &=& \left(\sum_{i=1}^{K} \vec{w}_i \mat{B}_i\right)^{-1} \vec{p},
\end{eqnarray}
where $\mat{B}_i$ is the transformation matrix for joints $i \in \{1 \hdots K\}$, and $\vec{w}_i$ its blend-weight. Eq.~\eqref{eq:can_trans} and~\eqref{eq:can_trans2} define the canonical mapping network ${f_\mathrm{can}:(\vec{p},\mathbf{P}_{\pose}) \mapsto \bar{\vec{p}}}$.

\subsection{Displacement Field Network}
\label{def_net}

A surface expressed implicitly as the zero level set ${\{\vec{x}\in\mathbb{R}^3\,|\, SDF(\vec{x})=0\}}$ can be deformed to obtain a new surface by applying transformations to the points themselves. By applying a displacement ${\vec{x}^\prime = \vec{x}+\Delta\vec{x}}$ we will effectively displace the zero level set ${\{\vec{x}\in\mathbb{R}^3 \,|\, SDF(\vec{x}^\prime)=0\}}$, and consequently the surface defined. 
To model subtle non-rigid deformations, we propose to explicitly learn a pose-dependent displacement field in the canonical space: 
\begin{equation}
\Delta \vec{\bar{p}_{\mathrm{\pose}}} = f_{\mathrm{def}}(\vec{\bar{p}}, \vec{w} \circ \pose, \mathbf{\bar{P}}_{\pose})    
\end{equation}
where $\vec{w} \circ \pose$ is weighted SMPL pose parameter and $\mathbf{\bar{P}}_{\pose}$ is pose encoding computed in the canonical space. We denote the space of pose encoding as ${(\vec{w} \circ \pose,\mathbf{\bar{P}}_{\pose}) \in \mathcal{P}}$.
Note that these are \emph{not per-vertex displacements}, but a \emph{learned continuous displacement field} $\mathbb{R}^3\mapsto \mathbb{R}^3$, and hence we are not restricted to a fixed mesh with fixed number of vertices.

\subsection{Canonical SDF and Normal Prediction}
\label{sdf_net}
Given points in canonical space and the pose  encoding $(\vec{w} \circ \pose,\mathbf{\bar{P}}_{\pose}) \in \mathcal{P}$, we learn a signed distance field in canonical space. 
The canonical SDF $(f_{\mathrm{CSDF}} : \mathbb{R}^3 \times \mathcal{P} \rightarrow \mathbb{R})$ is parameterized with a fully-connected neural network. As seen in Fig.~\ref{fig:method}, the network predicts SDF values for points in the canonical space with the following equation:

\begin{equation}
\label{eq:sdf_pred}
  d^{*} =  f_{\mathrm{CSDF}}(\vec{\bar{p}} + \Delta \vec{\bar{p}_{\mathrm{\pose}}},\vec{w} \circ \pose, \mathbf{\bar{P}}_{\pose})
\end{equation}
where the canonical point $\bar{\vec{p}}=f_\mathrm{can}(\vec{p},\mathbf{P}_{\pose})$, and the displacement $\Delta \vec{\bar{p}}= f_{\mathrm{def}}(\vec{\bar{p}}, \vec{w} \circ \pose, \mathbf{\bar{P}}_{\pose})$ are computed with the canonical mapping and displacement field networks respectively.
This formulation has several nice properties. First, the network can implicitly learn a canonical shape of clothing/human, without ever explicitly un-posing the scans. Note that if the motion consists exclusively of only articulation and displacement, then we could directly learn $f_{\mathrm{CSDF}}(\vec{\bar{p}} + \Delta \vec{\bar{p}})$ without further conditioning (in practice we condition on pose to give the network the flexibility to learn shapes which can not be modeled via displacements from a canonical shape).
Second, we can supervise Eq.~\eqref{eq:sdf_pred} with distances computed in the original posed scans.  

In addition to signed distances, we predict the normals for points near the surface, to generate more realistic looking results. We predict normals in canonical space, using $f_{\mathrm{norm}}$ and then transform it to pose space using blend weights, computed from Eq.~\eqref{eq:can_trans} and  $\mat{B}_i$: 
\begin{equation}
\label{eq:normal_pred}
\vec{n^{*}} =   \sum_{i=1}^{K} \vec{w}_i \mat{B}_i  f_{\mathrm{norm}}(\vec{\bar{p}} + \Delta \vec{\bar{p}_{\mathrm{\pose}}}, \vec{w} \circ \pose, \mathbf{\bar{P}}_{\pose}  )
\end{equation}

\subsection{Multi-subject model using Neural-GIF}
\label{multi_subject}
We extend our proposed framework for multi-subject setting, \ie we learn multiple shapes in one single model. For this we take motivation from SMPL~\cite{SMPL:2015} model, where per-vertex shape dependent displacement is added and extend the idea to a continuous shape dependent displacement field. To incorporate shape dependent displacement field, we introduce another displacement network($f_{\mathrm{def-\shape}}$), which predicts continuous displacement field in $\mathbb{R}^3$, given shape parameter($\shape$) as input:
\begin{equation}
\Delta \vec{\bar{p}_{\mathrm{\shape}}} = f_{\mathrm{def-\shape}}(\vec{\bar{p}}, \shape)    
\end{equation}
We further modify our model by taking SMPL shape($\beta$) as additional input to each sub-network and adding $\Delta \vec{\bar{p}_{\mathrm{\shape}}}$ to $\vec{\bar{p}} + \Delta \vec{\bar{p}_{\mathrm{\pose}}}$ before applying $f_{\mathrm{CSDF}}(*)$ and $f_{\mathrm{norm}}(*)$.
\begin{figure*}[t]
	\centering
	\includegraphics[width=0.48\textwidth]{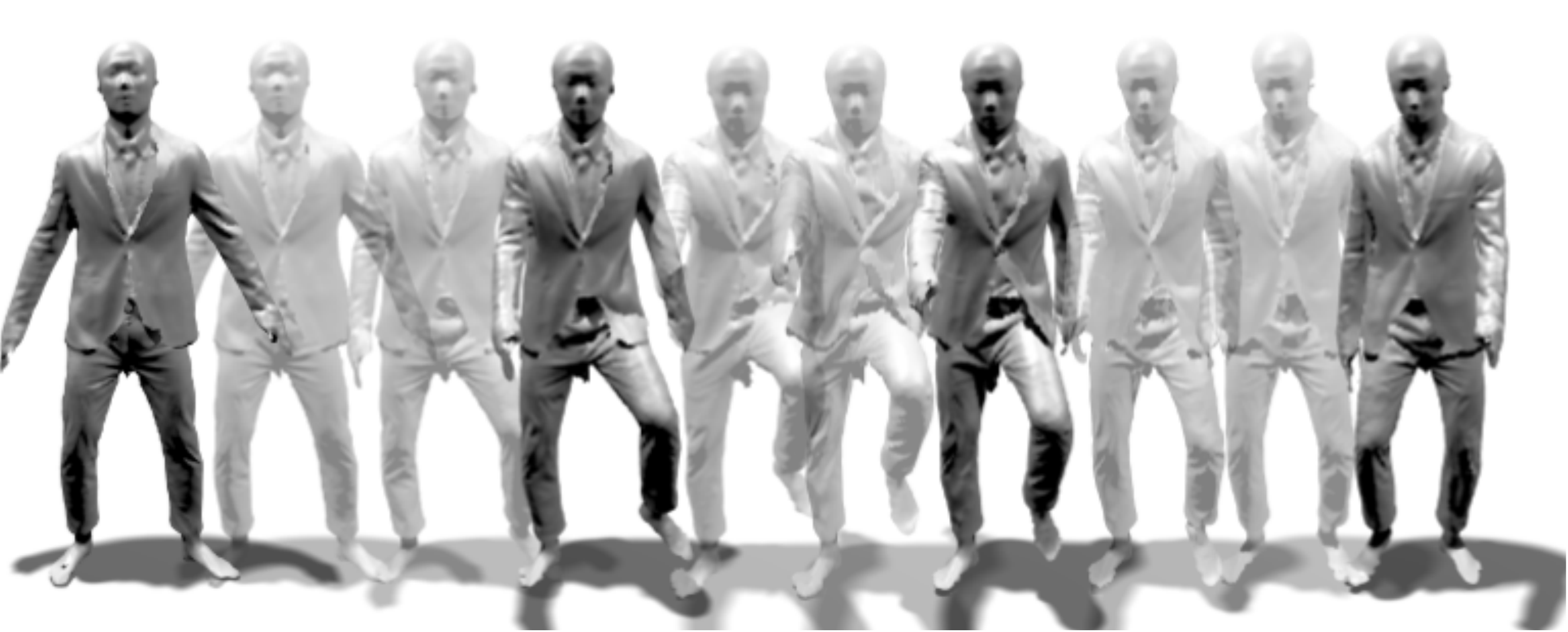}
	\includegraphics[width=0.48\textwidth]{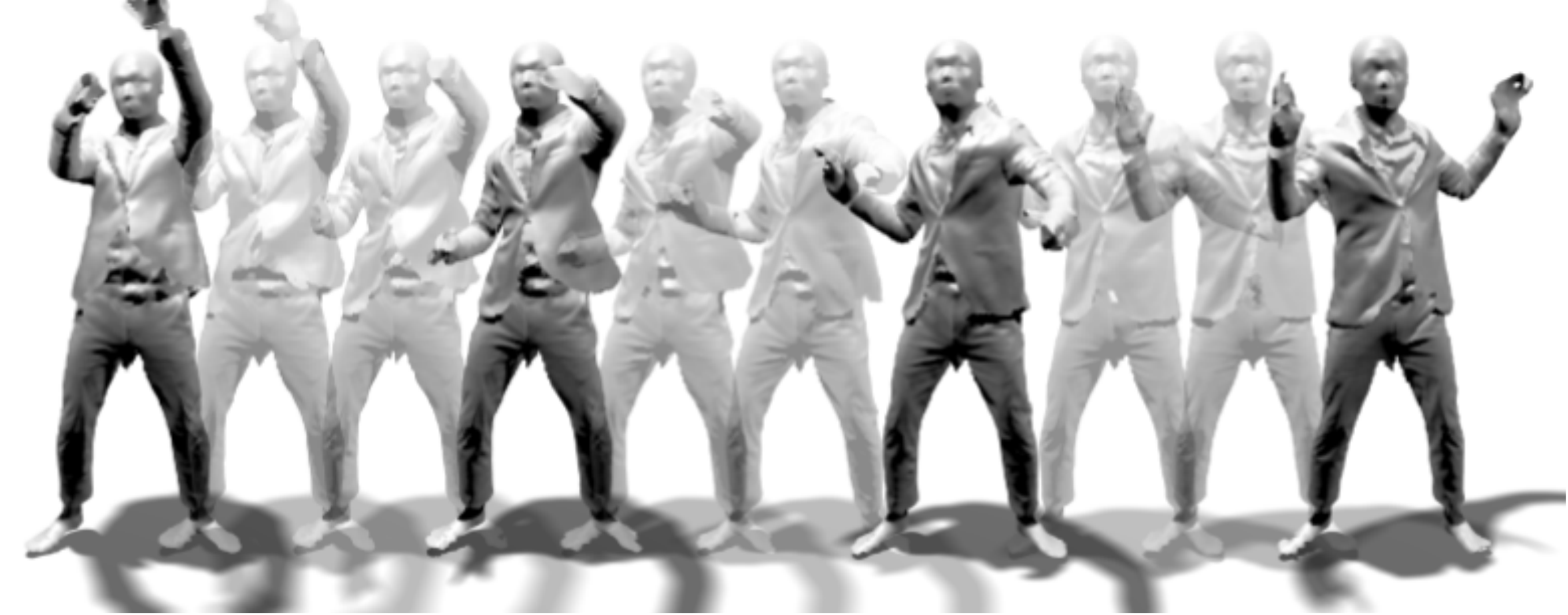}
	\caption{\textbf{Re-animating CAPE:} Reposing a character from CAPE dataset~\cite{ma20autoenclother} using our proposed method. We show results on walking motion sequence (left) and on playing volleyball motion sequence (right).}
	\label{fig:cape_results}
	\vspace{-0.2cm}
\end{figure*}

\subsection{Training and Inference}
\label{training}
\noindent\textbf{Training:} %
Each of the sub-networks is implemented using an MLP. Following~\cite{mildenhall2020nerf}, we use a positional encoding for query point $\vec{p}$. We follow a multi-stage training regime for faster convergence and stable training. First we pre-train the canonical mapping network ($f_{\mathrm{can}}$) using SMPL meshes generated from the training motion sequences and initially supervise blend weights with ground truth (only known at the SMPL surface). Then we train the canonical mapping network ($f_{\mathrm{can}}$) and the canonical SDF ($f_{\mathrm{CSDF}}$) end-to-end. Then in the final step, we freeze canonical mapping network ($f_{\mathrm{can}}$) and train $( \,f_{\mathrm{CSDF}}, \, f_{\mathrm{def}}) $ together end-to-end. For training the normal prediction network, we train only for points which are near the surface, \emph{i.e.} $| d^{*} | < \delta $, once the loss of $f_{\mathrm{CSDF}}$ has stabilized. We train our network using the following point-wise losses:
\begin{equation}
\label{eq:loss}
 \loss_{\mathrm{SDF}} = ||d^{*} - d||_{2} \;  \; \;  \;  \;  \; \loss_{\mathrm{norm}} = 1 - \vec{n}^T \vec{n}^*
\end{equation}
where $d^{*}, d$ represent the predicted and ground truth signed distances respectively, and $\vec{n}^{*}, \vec{n}$ denote the predicted and ground truth normals for a point in the posed space.
We also added a regularizer for pose dependent displacement field network, given by:
\begin{equation}
\label{eq:loss_reg}
 \loss_{\mathrm{\Delta \vec{\bar{p}_{\mathrm{\pose}}}}} = ||\Delta \vec{\bar{p}_{\mathrm{\pose}}}||_{2}^2 
\end{equation}

We generate training data by sampling points on the surface of the ground-truth mesh and randomly displace these points with noise $\sigma= \{ 0.01, \, 0.1 \}$. Similar to~\cite{chibane2020ndf}, the ground-truth signed distances at these samples are computed by casting randomized rays and checking the parity~\cite{Jacobson-13-winding}. We use $K=24$ number of joints, for fully body meshes and follow~\cite{patel2020} for skirt and shirt meshes.

\noindent\textbf{Inference:} For generating meshes from our method, we first sample points in space and predict $d^{*}$, using Eq.~\eqref{eq:sdf_pred}. We reconstruct our meshes using marching cubes~\cite{Lorensen87marchingcubes:} on the predicted SDF. Our reconstructed meshes are at 256 voxel resolution. Then we predict normals for the vertices on the reconstructed surface, using $f_{\mathrm{norm}}$. Note that unlike~\cite{Saito:CVPR:2021}, we directly query our point in pose space, so there is no need for extra re-posing step.

\section{Experiments and Results}
Our proposed method can be used to learn non-rigid pose-dependent deformations in human/clothing meshes. We show results of our method on various datasets in Sec.~\ref{re-animscans} and compare with ~\cite{deng2019neural,Saito:CVPR:2021} in Sec.~\ref{comparison}. We also evaluate our model in multiple shape setting in Sec.~\ref{multi_shape}.

\noindent\textbf{Datasets}: For evaluation of single-shape and clothing model, we use CAPE dataset~\cite{ma20autoenclother,ponsmoll2017clothcap}, DFAUST~\cite{dfaust, pons2015dyna}, TailorNet~\cite{patel2020} and a few human scan sequences with complex clothing that we captured in our stage, which we will refer to as \frl. 
For multiple subject setup, we evaluate our method on three datasets: 1) MoVi~\cite{Ghorbani_2021,AMASS:ICCV:2019}, 2) SMPL pose and shape dataset and 3) DFAUST registrations~\cite{dfaust, pons2015dyna} and compare with~\cite{LEAP:CVPR:21}. For SMPL pose and shape dataset, we use the 890 poses provided in SMPL model and create 9 different body shapes, by changing first 3 principal components of shape-space. For the CAPE and TailorNet, the SMPL parameters are available in the dataset. For others, we fit SMPL~\cite{bhatnagar2020ipnet, bhatnagar2020loopreg} to obtain the pose and shape.

We split the dataset, such that the test set contains 2 unseen motion sequences for CAPE and DFAUST. For TailorNet and SMPL pose-shape dataset, we follow the train/test split used in~\cite{patel2020}. For MoVi, we follow the train/test split used in~\cite{LEAP:CVPR:21}.
In order to show that Neural-GIF can model complex geometries, loose clothing and topolgies, we test our method on the \frl\ which is a dataset of complex clothing like jacket, loose shrugs etc. 
\begin{figure*}[t]
	\centering
	\includegraphics[width=0.32\textwidth]{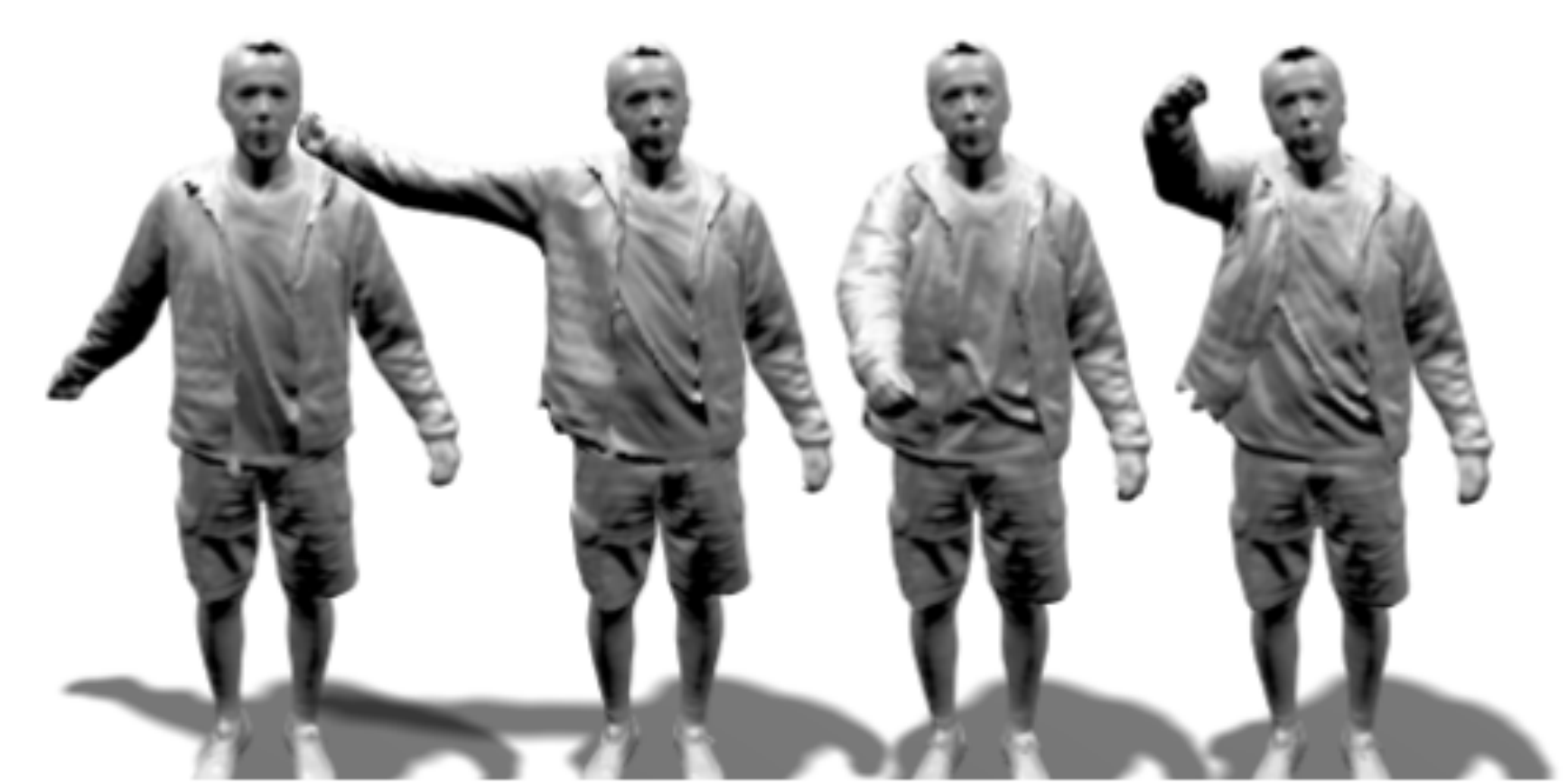}
	\includegraphics[width=0.32\textwidth]{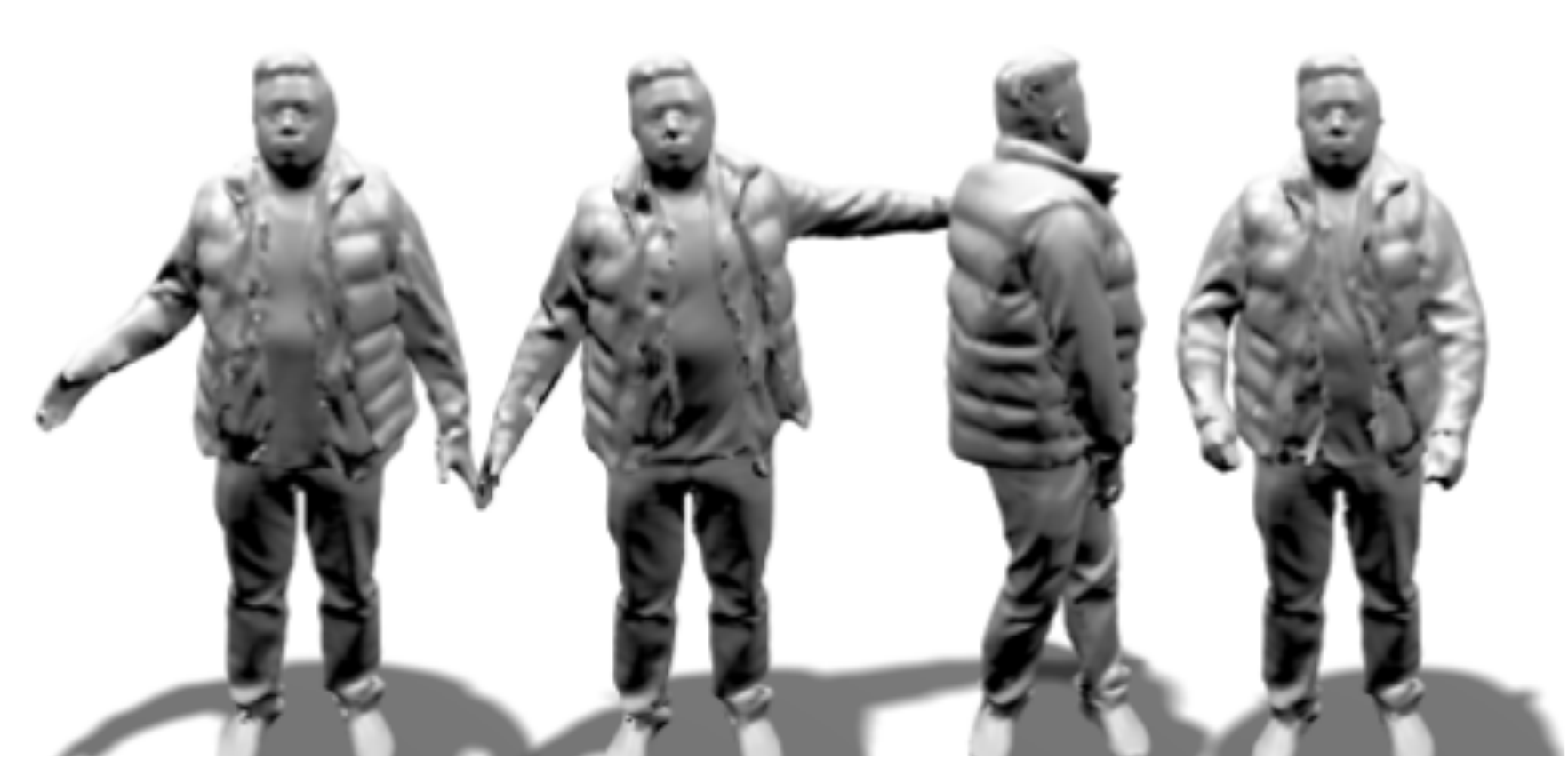}
	\includegraphics[width=0.32\textwidth]{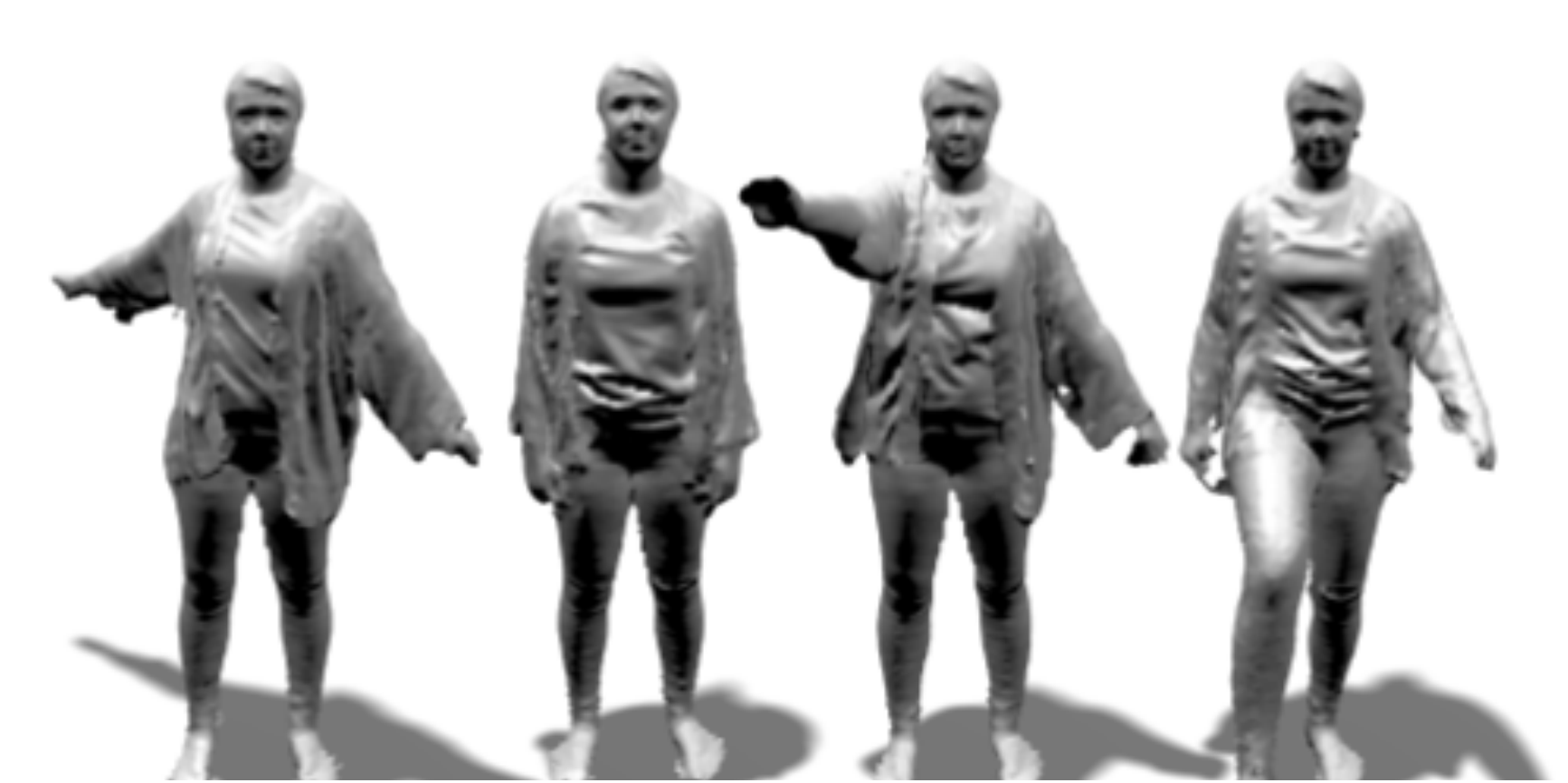}
	\caption{\textbf{Modeling people in complex clothes:} Results of Neural-GIF for subjects from the \frl\ dataset. Neural-GIF  can model well complex cloth geometries, like hoodies (left), puffer jacket (middle) and loose clothes like shrugs (right).}
	\label{fig:xyz_results}
\end{figure*}

\subsection{Reanimating people in clothing}
\label{re-animscans}
\noindent\textbf{Modeling people in complex clothing:} We use our Neural-GIF, to animate characters in new poses. We generate results for various characters of the \frl\ and CAPE datasets. We provide results on poses of the test set (volleyball and walking sequences) of the CAPE dataset in Fig.~\ref{fig:cape_results} and on the \frl\ dataset in Fig.~\ref{fig:xyz_results}. Our method generates accurate reconstructions and generalizes to unseen poses, while preserving the structure of identity and generating fine details. Since we directly learn from scans, which are noisy, \eg, near the hands, feet and inner-thigh, the reconstruction around these areas tends to be a bit noisy and unstructured. Modeling such types of clothing using template-based methods is not straightforward due to the open-layered structure of the clothes on top of the body. As a result, a single template cannot explain two extreme deformations of a jacket, (\ie, on one end when a layer of the jacket is close to the body surface and on the other side of the spectrum when the layer is far from the body surface due to some pose like lifting hands in the air).
Our method produces large deformations for loose garments like open jackets and shrugs(Fig.~\ref{fig:xyz_results}) as opposed to template-based methods. This is possible because we do not rely on blend weights associated to the underlying body, but instead we predict blend weights for every point in the space as a function of pose, for a given subject. So our network learns to associate each point correctly to joints \eg a point on the surface of jacket near torso will have different association (weight) in the rest pose and slightly different association when the person is lifting hands. Our canonical mapping network learns this pose-dependent association from data. 

\begin{figure}[t]
	\centering
	\begin{overpic}[width=0.24\textwidth,unit=1mm]{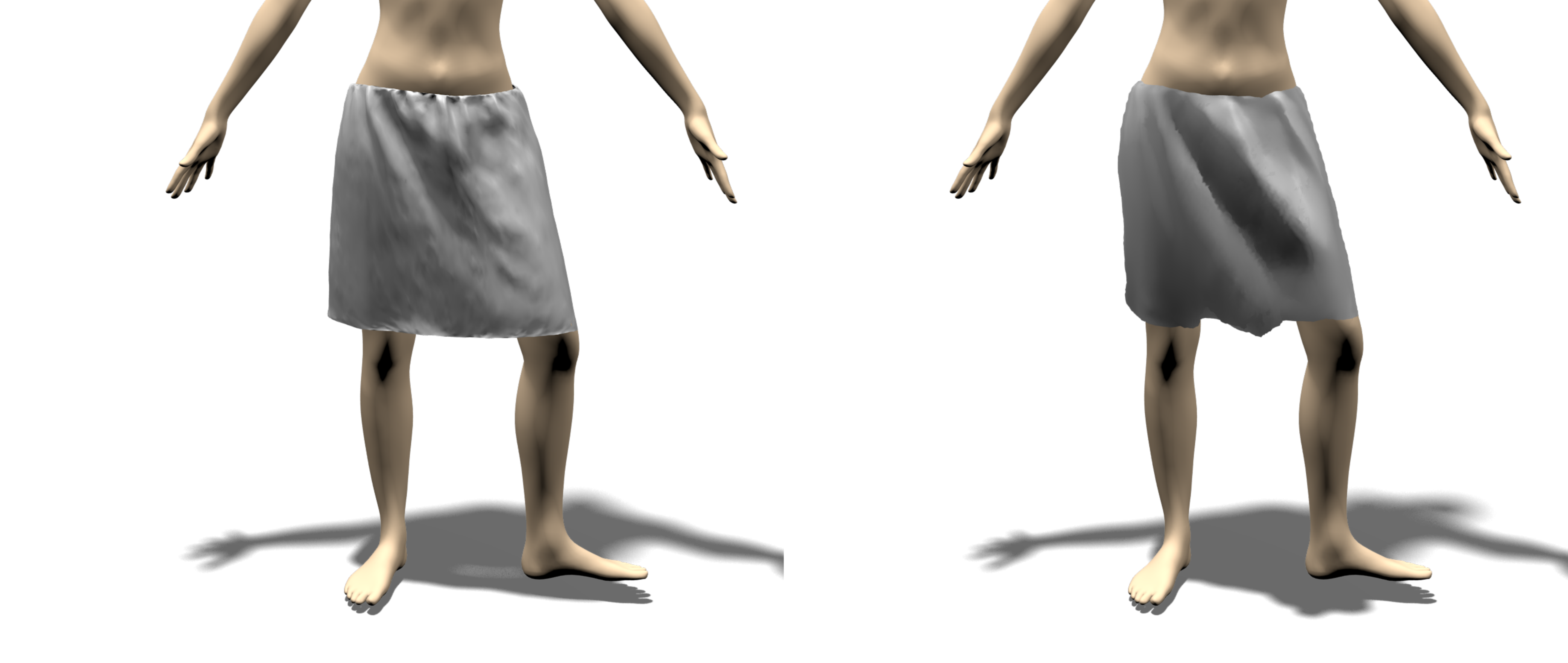}

\end{overpic}	\begin{overpic}[width=0.24\textwidth,unit=1mm]{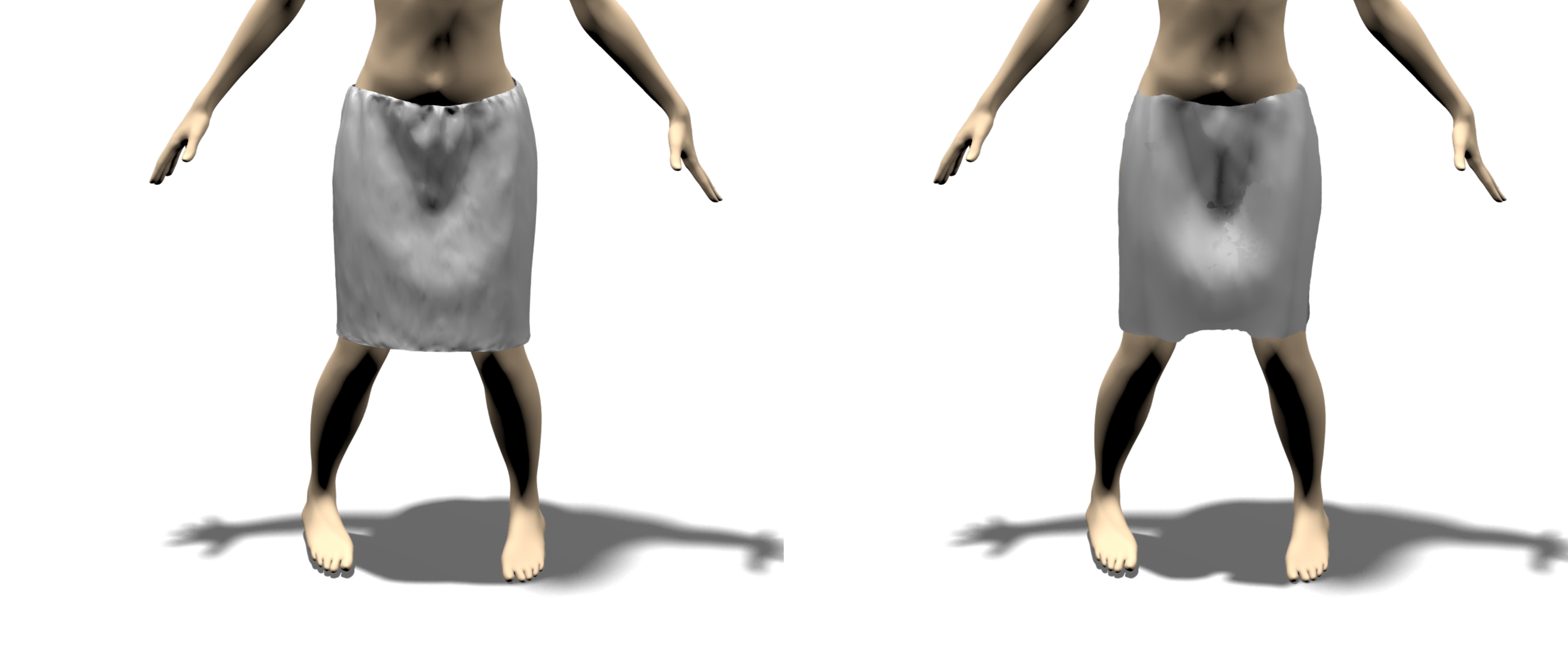}

\end{overpic}
	\begin{overpic}[width=0.24\textwidth,unit=1mm]{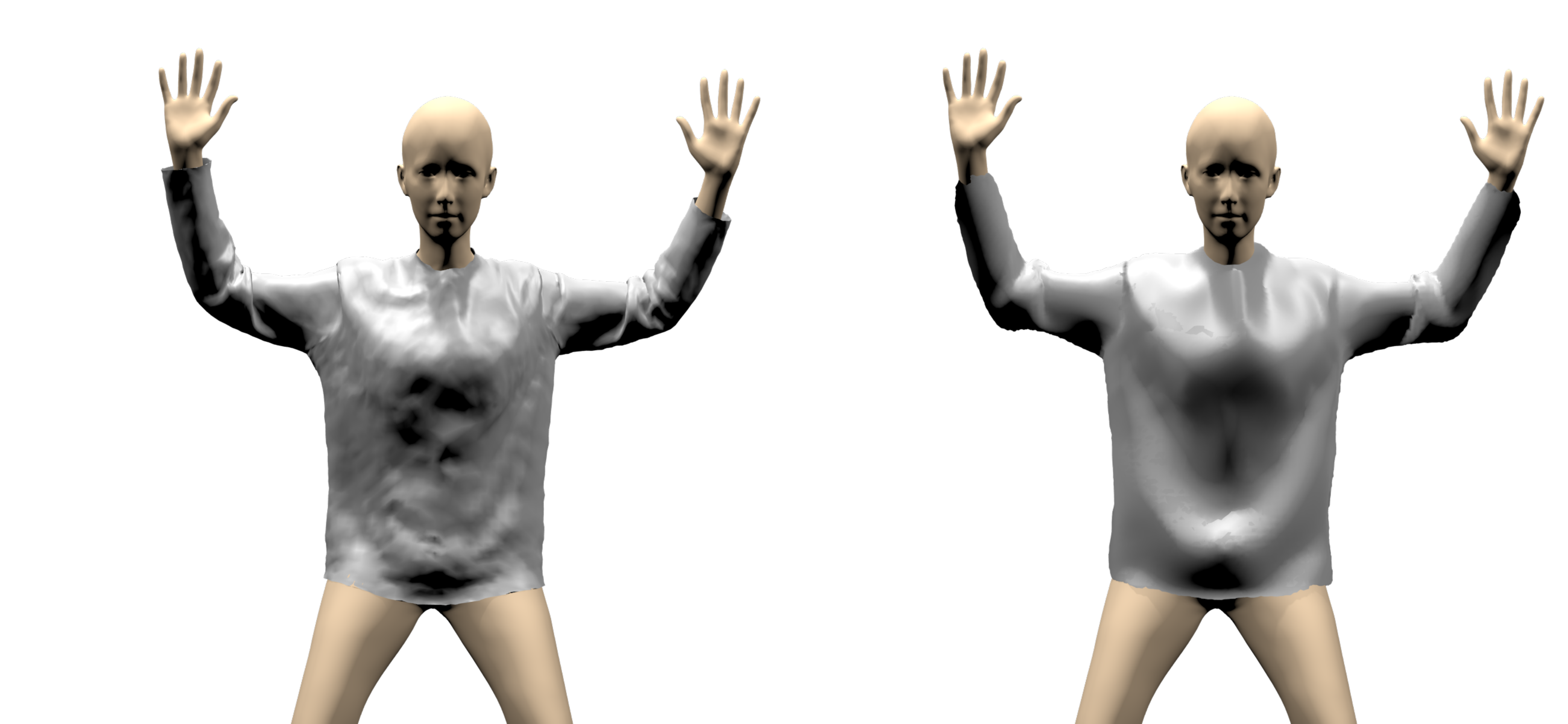}
	\put(5,0){\colorbox{white}{\parbox{0.4\linewidth}{%
    TailorNet }}}
    	\put(27,0){\colorbox{white}{\parbox{0.4\linewidth}{%
    $\; \;$Ours }}}

\end{overpic}	\begin{overpic}[width=0.24\textwidth,unit=1mm]{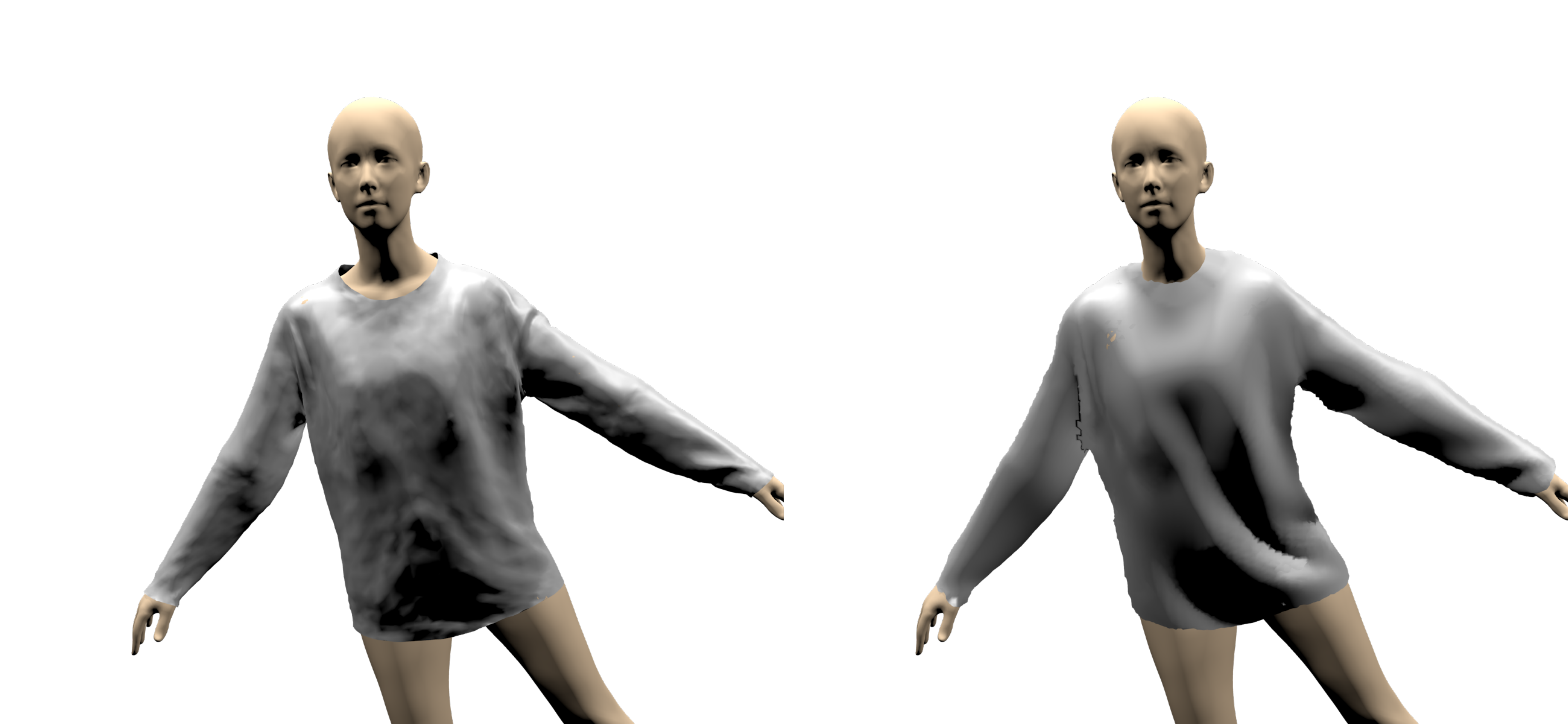}
	\put(5,0){\colorbox{white}{\parbox{0.4\linewidth}{%
    TailorNet }}}
    	\put(25,0){\colorbox{white}{\parbox{0.4\linewidth}{%
    $\; \;$Ours }}}

\end{overpic}

	\caption{\textbf{Comparison with TailorNet:} Qualitative comparison on skirt(top) and shirt(bottom) meshes from the TailorNet dataset.} 
	\label{fig:tailornet}
\end{figure}

\noindent\textbf{Modeling separate clothing items:} Since our method is not limited by a pre-defined template or topology, we can also train Neural-GIF on the clothing meshes of shirts and skirts from the TailorNet dataset, and for completeness, also compare to TailorNet~\cite{patel2020}. 
It is important to note that such a comparison is not totally fair for our method. First, the training meshes are of a fixed resolution dictated by the template used in TailorNet, so our model is upper-bounded by this. Second, registration is noise free because meshes are obtained from physics simulations (ideal case for template-based methods) -- this allows the mapping of every shape to a vector, which makes learning significantly easier for TailorNet -- correspondence is known for TailorNet whereas is not known for Neural-GIF. 
Despite this, qualitatively Neural-GIF is comparable to TailorNet~\cite{patel2020} as seen in Fig.~\ref{fig:tailornet}. We can see that our method generalizes to new unseen poses and predicts fine details and deformations. Quantitatively we find that point to surface distance for TailorNet results is slightly better than our method. We produce $11.14 \,$mm and $10.98 \,$ mm error for skirt and shirt clothing, while TailorNet's results are $9.82 \,$mm and $7.28 \,$ mm respectively.
Another source of slight inaccuracy originates from our mesh reconstruction method for open surfaces. Since shirt/skirts are a thin layered surface, we reconstruct the mesh using a small threshold of 0.01mm in marching cubes, which might result in slightly thicker surfaces, and could be addressed by switching to unsigned neural distance fields~\cite{chibane2020ndf}. 
This comparison shows that when registration is possible (easy clothing animated with physics simulation) it helps learning.
\subsection{Modeling soft-tissue dynamics}
\label{soft-tissue}
\begin{figure*}[t]
  \centering
  	\begin{overpic}[width=0.33\textwidth,unit=1mm]{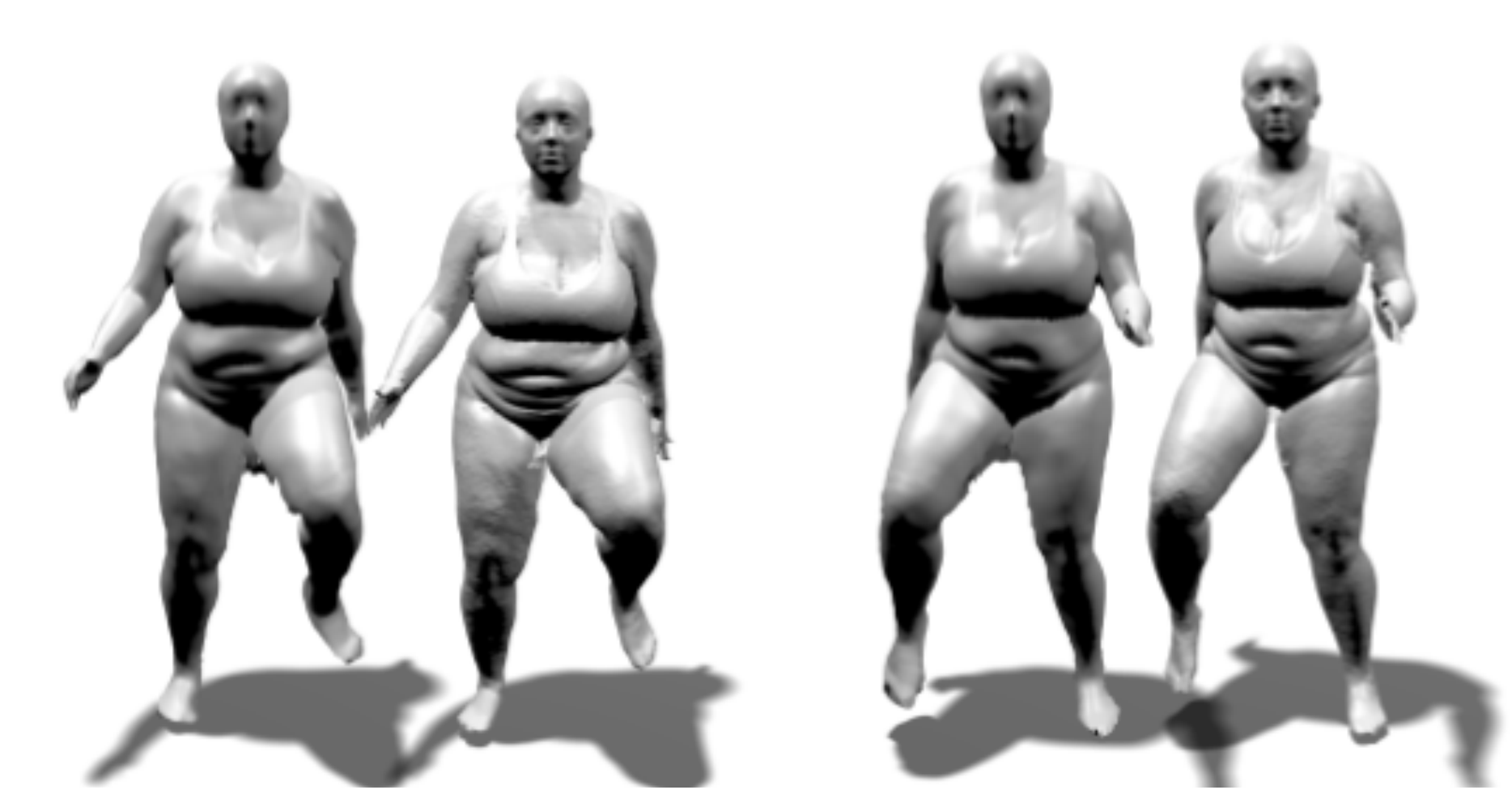}
	\put(5,30){\colorbox{white}{\parbox{0.4\linewidth}{%
     Ours}}}
    	\put(18,30){\colorbox{white}{\parbox{0.4\linewidth}{%
     GT}}}
	\put(35,30){\colorbox{white}{\parbox{0.4\linewidth}{%
     Ours}}}
    	\put(45,30){\colorbox{white}{\parbox{0.4\linewidth}{%
     GT}}}

\end{overpic}
 	\begin{overpic}[width=0.33\textwidth,unit=1mm]{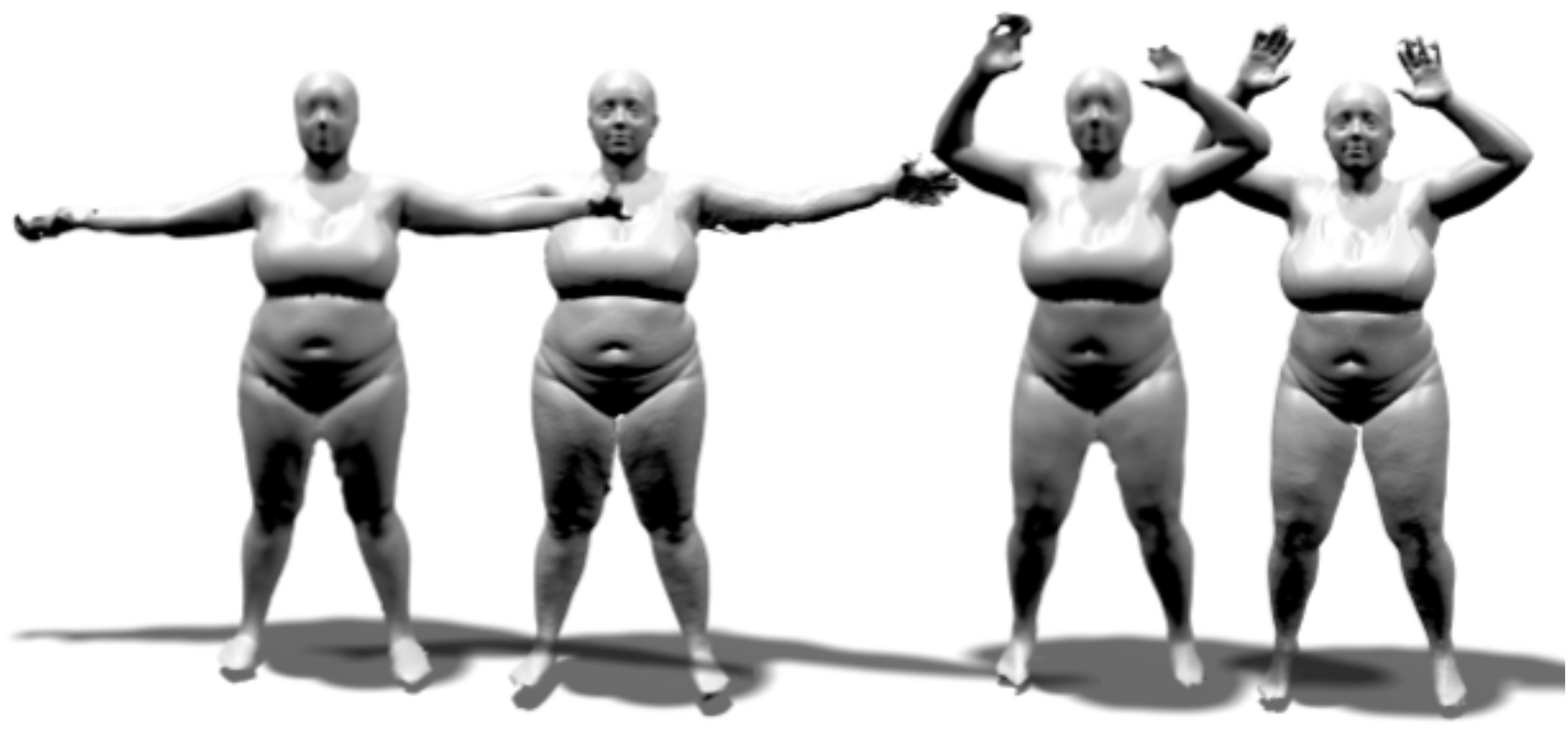}

 	\put(10,30){\colorbox{white}{\parbox{0.4\linewidth}{%
     Ours}}}
    	\put(20,30){\colorbox{white}{\parbox{0.4\linewidth}{%
     GT}}}
	\put(35,30){\colorbox{white}{\parbox{0.4\linewidth}{%
     Ours}}}
    	\put(45,30){\colorbox{white}{\parbox{0.4\linewidth}{%
     GT}}}

\end{overpic}
   	\begin{overpic}[width=0.33\textwidth,unit=1mm]{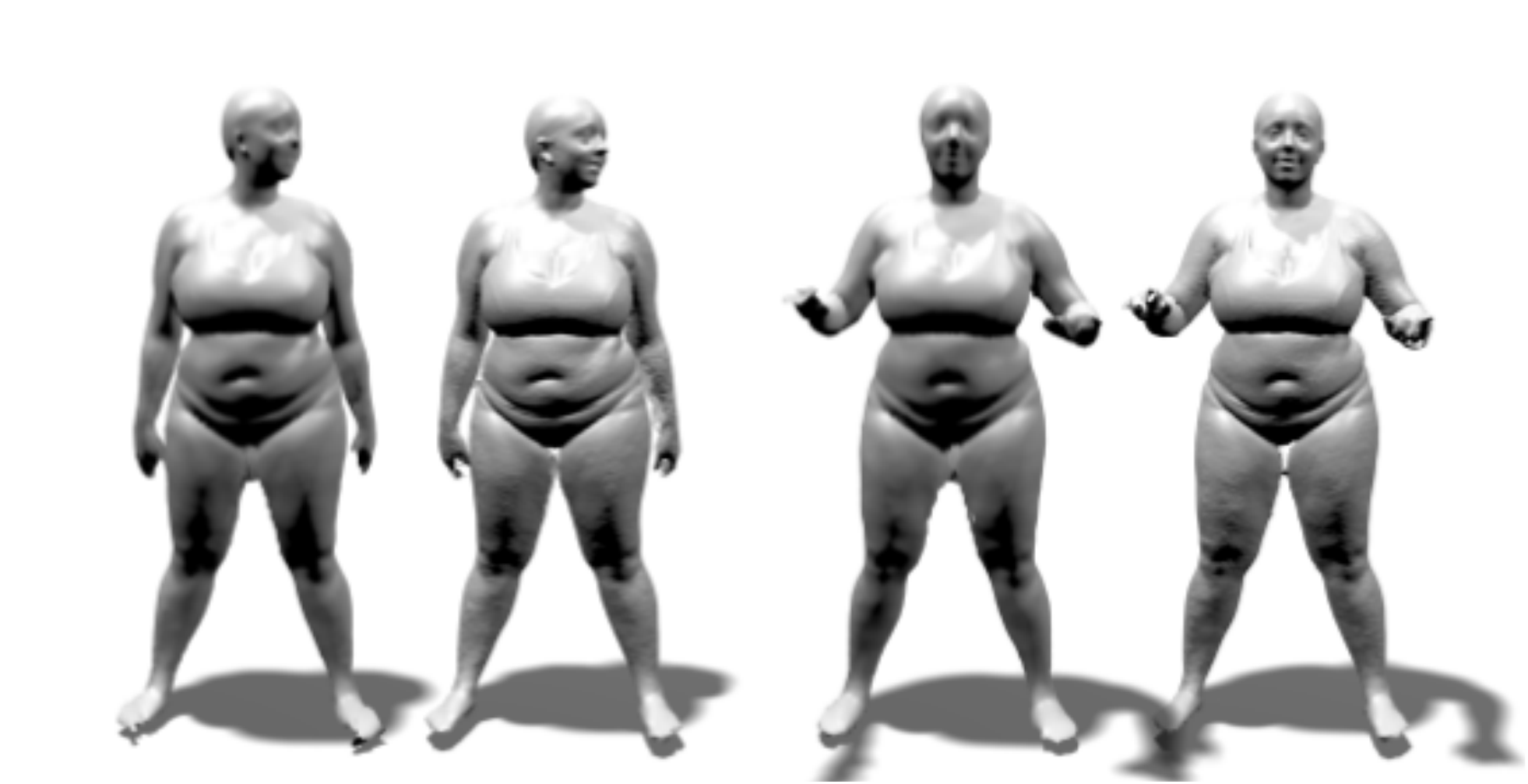}
	\put(5,30){\colorbox{white}{\parbox{0.4\linewidth}{%
     Ours}}}
    	\put(18,30){\colorbox{white}{\parbox{0.4\linewidth}{%
     GT}}}
	\put(32,30){\colorbox{white}{\parbox{0.4\linewidth}{%
     Ours}}}
    	\put(45,30){\colorbox{white}{\parbox{0.4\linewidth}{%
     GT}}}

\end{overpic}

\caption{\textbf{Results on DFAUST~\cite{dfaust:CVPR:2017}: }We show our results on three different motion sequences (running, jiggle on toes, shake hips) from DFAUST and compare with ground-truth DFAUST data and visualize results in (prediction, GT) pairs. Neural-GIF is able to retain pose-dependent soft-tissue deformations present in the ground truth scans.} 
\label{fig:dyna_res}
\vspace{-0.2cm}
\end{figure*}

Neural-GIF models high frequency pose-dependent deformations for various shapes like full clothing, jackets and skirts. We now evaluate how well this representation can model soft-tissue dynamics, which are highly non-linear and high-order pose-dependent deformations. Thus, we evaluate our method on this task using the DFAUST~\cite{dfaust:CVPR:2017} dataset. We show our results on DFAUST~\cite{dfaust:CVPR:2017} in Fig.~\ref{fig:dyna_res}. We evaluate our method on a single subject with 14 sequences, which demonstrate extreme soft-tissue dynamics. %
In Fig.~\ref{fig:dyna_res} we provide results for 3 sequences and
observe that the network predicts soft-tissue dynamics/deformations. Our normal prediction network also helps in producing more realistic renders. We notice that since deformation in the facial region is not controlled by any keypoint and also the subjects' facial expressions change over time, we tend to produce smooth results around the face. 

\begin{table*}[t]
\centering
\caption{We quantitatively compare the results of our method with NASA~\cite{deng2019neural} and SCANimate~\cite{Saito:CVPR:2021}. We report point to surface distance (in mm) and IoU and F-Scores(\%) for comparison.}
\resizebox{\textwidth}{!}{
\begin{tabular}{lccccccccccc}
\toprule
\multirow{2}{*}{\diagbox{Dataset}{Model}} & \multicolumn{3}{c}{ NASA~\cite{deng2019neural}} &  \multicolumn{3}{c}{SCANimate~\cite{Saito:CVPR:2021}}& \multicolumn{3}{c}{\textbf{Ours (Neural-GIF)}}\\
\cmidrule(r){2-4}\cmidrule(r){5-7}\cmidrule(r){8-10}
\multicolumn{1}{c}{} & Point2Surface $\downarrow$ & IoU $\uparrow$ & F-Score $\uparrow$ & Point2Surface $\downarrow$ & IoU $\uparrow$  & F-Score $\uparrow$ & Point2Surface $\downarrow$ & IoU $\uparrow$  & F-Score $\uparrow$ \\
\midrule
CAPE~\cite{ma20autoenclother}         & 10.67  & 0.918 & 94.32 &  \textbf{5.82} & \textbf{0.957} & \textbf{98.51} & \textbf{5.86}   & \textbf{0.957} & \textbf{98.53} \\
\frl         & 23.26  & 0.780 & 57.29   & 7.32 & 0.953 & 97.32 & \textbf{4.73}  & \textbf{0.967} & \textbf{99.15} \\
DFAUST~\cite{dfaust:CVPR:2017}         & 10.52  & 0.939 & 95.48  & 3.79 & 0.971 & 99.50 & \textbf{3.21}  & \textbf{0.972}& \textbf{99.56} \\
\bottomrule
\end{tabular}
\label{tab:quant}
}
\end{table*}

\subsection{Comparisons and Ablation Studies}
\label{comparison}
\begin{figure}[t]
	\centering
   	\begin{overpic}[width=0.5\textwidth,unit=1mm]{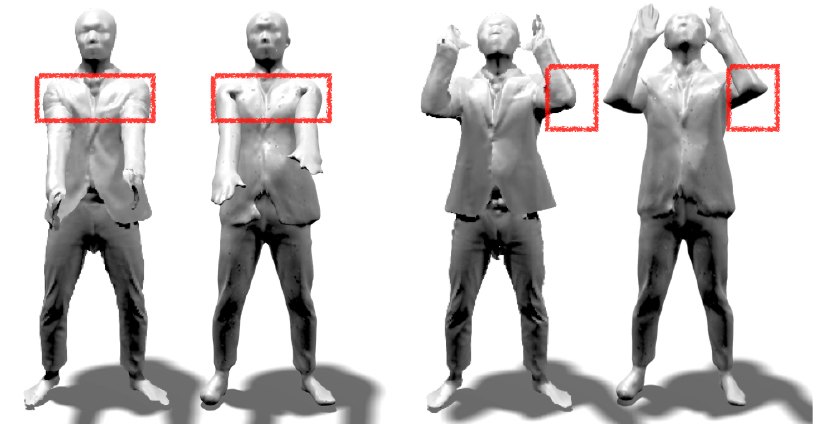}
	\put(5,0){\colorbox{white}{\parbox{0.4\linewidth}{%
     Ours}}}
    	\put(20,0){\colorbox{white}{\parbox{0.4\linewidth}{%
     SCANimate}}}
	\put(45,0){\colorbox{white}{\parbox{0.4\linewidth}{%
     Ours}}}
    	\put(60,0){\colorbox{white}{\parbox{0.4\linewidth}{%
     SCANimate}}}

\end{overpic}
	\caption{\textbf{Comparison with SCANimate:} We compare the results of our method on the CAPE dataset with SCANimate~\cite{Saito:CVPR:2021}. Our model preserves more details and does not have posing artifacts.}
	\label{fig:scan_comp}
\end{figure}

\begin{figure}[t]
	\centering
	\begin{overpic}[width=0.48\textwidth,unit=1mm]{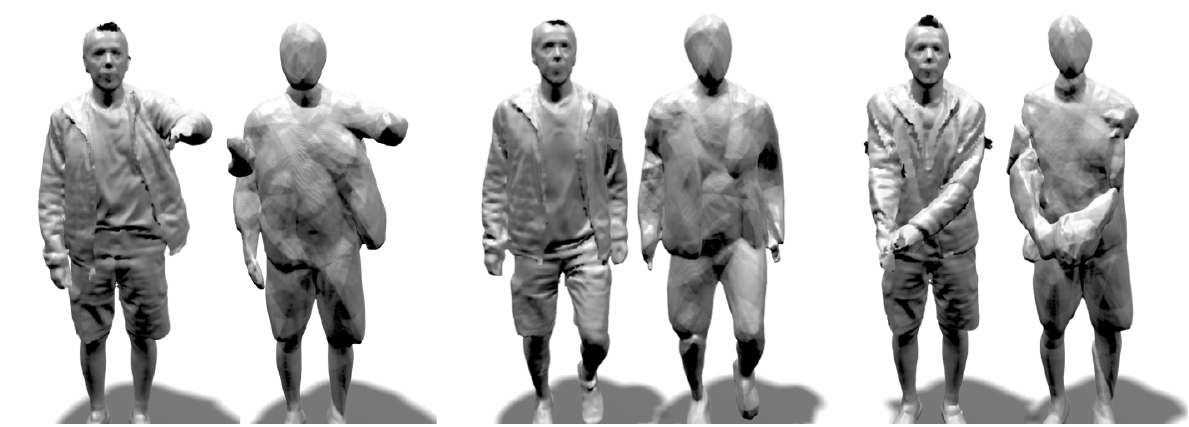}
	\put(4,0){\colorbox{white}{\parbox{0.4\linewidth}{%
     Ours}}}
    	\put(17,0){\colorbox{white}{\parbox{0.4\linewidth}{%
     NASA}}}
	\put(35,0){\colorbox{white}{\parbox{0.4\linewidth}{%
     Ours}}}
    	\put(46,0){\colorbox{white}{\parbox{0.4\linewidth}{%
     NASA}}}
	\put(60,0){\colorbox{white}{\parbox{0.4\linewidth}{%
     Ours}}}
    	\put(72,0){\colorbox{white}{\parbox{0.4\linewidth}{%
     NASA}}}
\end{overpic}

	\caption{\textbf{Comparison with NASA:} We compare the results of our method (bottom) on the \frl\ dataset with NASA~\cite{deng2019neural} (top). Our model reconstructs complex clothing without any part-based artifacts and generates realistic pose-dependent deformations.}
	\label{fig:nasa_com}
\end{figure}

\noindent\textbf{Comparison with SoTA:} For the single shape model, we compare our method with~\cite{deng2019neural, Saito:CVPR:2021} and provide the quantitative evaluations in Table~\ref{tab:quant}. In NASA~\cite{deng2019neural} pose-dependent occupancy is represented using part-based implicit functions. NASA produces good results for human meshes from~\cite{dfaust:CVPR:2017}, however it cannot model fine details and complex or loose clothing like jackets. We observe from Fig.~\ref{fig:nasa_com}, that NASA generates rigid and overly smooth results, compared to our model. NASA also suffers from part-based artifacts, such as intersecting body parts in poses like bending elbows or knees. This problem is even more prominent if there are significantly less data points for such poses. Since we learn the whole body shape using a single network our approach does not encounter such issues. From Tab.~\ref{tab:quant} we observe a significant drop in terms of IoU on \frl\ when using NASA, which is because the network cannot learn the loose deformations of jackets and hoodies, as one can observe in Fig.~\ref{fig:nasa_com}. Quantitatively, SCANimate~\cite{Saito:CVPR:2021} performs better than our method on CAPE, as seen in Table~\ref{tab:quant}, first row, but we notice that results of SCANimate are less detailed and have posing artifacts due to LBS, as highlighted in figure~\ref{fig:scan_comp}.
We also show significant improvements in terms of the quality of details produced by our method. We accredit this to the combination of the displacement field network, the normal prediction and positional encoding of query point.%

\begin{table}[t]
\centering
\caption{\textbf{Ablation}: Quantitative evaluation between nearest neighbour based SMPL weights and learned weights from canonical mapping network.}
\resizebox{0.5\textwidth}{!}{
\begin{tabular}{lccccc}
\toprule
\multirow{2}{*}{\diagbox{Dataset}{Model}} & \multicolumn{2}{c}{SMPL weights} & \multicolumn{2}{c}{Canonical mapping network}\\
\cmidrule(r){2-3}\cmidrule(r){4-5}
\multicolumn{1}{c}{} & Point2Surface $\downarrow$ & IoU $\uparrow$ & Point2Surface $\downarrow$ & IoU $\uparrow$  \\
\midrule
CAPE~\cite{ma20autoenclother}  & 12.93  & 0.866  &  7.93 &  0.955  \\
\frl        & 11.89 &  0.874 & 9.18 & 0.963  \\
DFAUST~\cite{dfaust:CVPR:2017}        & 18.38  & 0.806 & 3.28 &  0.972  \\

\bottomrule
\end{tabular}
\label{tab:ablation}
}
\end{table}

\begin{figure}[t]
	\centering
	   	\begin{overpic}[width=0.48\textwidth,unit=1mm]{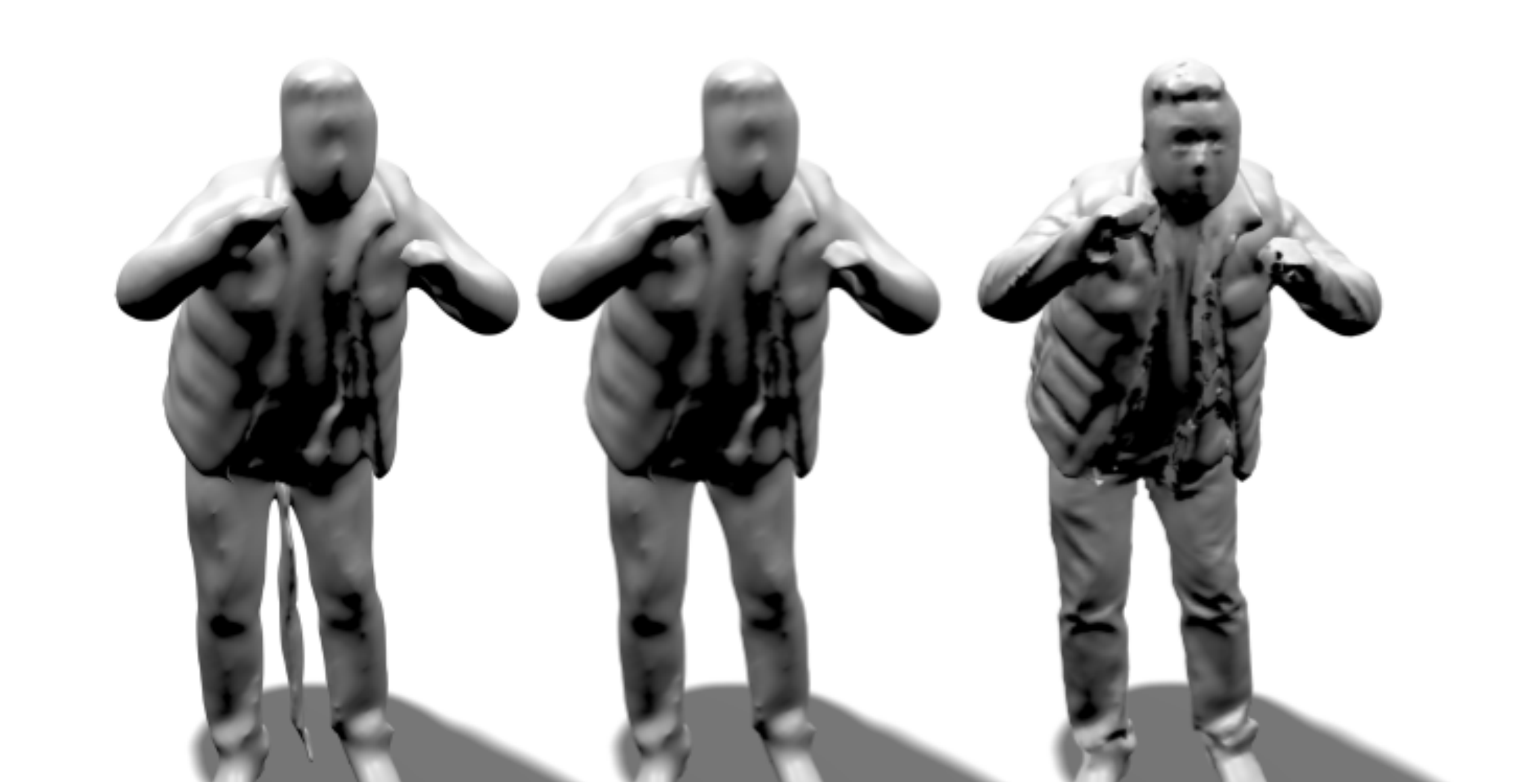}	

 	\put(10,44){{$f_{\mathrm{CSDF}}$+ }}
 	\put(5,40){{ SMPL weights}}
 	\put(33,42){{$f_{\mathrm{CSDF}}$+ $f_{\mathrm{can}}$}}

	\put(63,44){{\parbox{0.16\textwidth}{%
     Ours}}}
     	\put(58,40){{\parbox{0.16\textwidth}{%
     (complete)}}}

\end{overpic}
	\caption{\textbf{Ablation:} Canonical SDF+SMPL weights results in wrong association and hence result in wrong occupancy (left). Our canonical mapping network solves this problem by learning blend weights from data, but produces smooth results (middle). Finally displacement field network and normal prediction network helps in generating small pose-dependent details (right).}
	\label{fig:our_variant}
	\vspace{+0.13cm}
\end{figure}

\noindent\textbf{Ablation studies:} We now evaluate the significance of each module in our proposed method and report the result of ablation in Table~\ref{tab:ablation} and our complete model in Table~\ref{tab:quant}(last column).
First, we simply use the Canonical SDF to learn the distance field in the canonical space. %
We find the blend weights for every point in space ($\vec{p}$), by associating the blend weight of nearest point on the SMPL body surface. We transform $\vec{p}$ to $\bar{\vec{p}}$ using Eq.~\eqref{eq:can_trans} and calculate the SDF using Eq.~\eqref{eq:sdf_pred}. In Fig.~\ref{fig:our_variant}(left), we observe that if we simply use nearest point based blend weights, we obtain wrong reconstructions (space between the legs). We observe that for this case (Table~\ref{tab:ablation}, col.1), the IoU is relatively low as compared to the other two versions of our model and also lower than NASA(see Table~\ref{tab:quant}) on DFAUST. This is because of incorrect body part associations, as shown in Fig.~\ref{fig:our_variant} and explained above. This problem is most prominent, near the inner thighs and especially when both the legs are not performing a symmetrical motion. The test split in DFAUST consists of jumping, running etc, the points between the legs randomly change their association between left and right legs and are mapped to a point in the canonical space, which may lie inside the canonical shape, for a given pose. 

We alleviate this problem, with our canonical mapping network to predict these blend weights (Table~\ref{tab:ablation}, col.2). The network learns correct associations based on data. Learning the SDF in the canonical space, simplifies the learning process, as the network learns a strong prior for the shape of the subject.
We observe from Fig.~\ref{fig:our_variant}(middle), that just using Canonical SDF + Canonical mapping network, does not produce fine pose-dependent deformations. Hence we introduce a displacement field network, to learn a pose-dependent displacement field in the canonical space for small non-rigid deformations and a normal prediction model. We observe that having a separate model to incorporate such fine details and training the network in a multi-step regime (Sec.~\ref{training}), helps to learn these deformations.  

\subsection{Multiple shape model}
\label{multi_shape}
As mentioned in section~\ref{multi_subject}, we extend our formulation to incorporate multiple body shapes in one model. We compare with concurrent method LEAP~\cite{LEAP:CVPR:21} and show the quantitative evaluation in Table~\ref{tab:multishape}. We model body shape variation as a continuous displacement field in canonical space. Even with a simple formulation, we perform comparably or better than LEAP. We also observe that our training is more stable than training LEAP. %

\begin{table}[t]
\centering
\caption{For multiple shape model we quantitatively compare the results of our method with LEAP~\cite{LEAP:CVPR:21}.}
\resizebox{0.5\textwidth}{!}{
\begin{tabular}{lccccc}
\toprule
\multirow{2}{*}{\diagbox{Dataset}{Model}} & \multicolumn{2}{c}{ LEAP~\cite{LEAP:CVPR:21}} & \multicolumn{2}{c}{Ours (Neural-GIF)}\\
\cmidrule(r){2-3}\cmidrule(r){4-5}
\multicolumn{1}{c}{} & Point2Surface $\downarrow$ & IoU $\uparrow$ & Point2Surface $\downarrow$ & IoU $\uparrow$  \\
\midrule
DFAUST~\cite{dfaust:CVPR:2017}  & 3.42  & 0.958 & \textbf{3.35} &  \textbf{0.963}  \\
MoVi~\cite{Ghorbani_2021}        & \textbf{3.19}  & \textbf{0.969} & 3.20 &  0.969  \\
SMPL        & 3.26  & 0.968 & \textbf{3.18} &  \textbf{0.971}  \\

\bottomrule
\end{tabular}
\label{tab:multishape}
}
\end{table}

\section{Conclusion}
We introduced Neural-GIF, a novel model to learn articulation and pose-dependent deformation for humans in complex clothing using an implicit 3D surface representation. Neural-GIF admits end-to-end learning directly from scans provided with their corresponding SMPL pose and shape parameters.  
The key idea of Neural-GIF is to express shape as displacement and articulation deformations of a canonical shape, a concept that has been widely used for parametric meshes, which we generalize here to neural implicit function learning. 
Neural-GIF can accurately model complex geometries of arbitrary topology and resolution, because our model does not require a pre-defined template, or non-rigid registration of a template to scans. 

We show significant improvements from prior work~\cite{deng2019neural, Saito:CVPR:2021}, in terms of robustness, ability to model complex clothing styles and retaining fine pose-dependent details. We believe that the use of generalized implicit functions with our canonical mapping and displacement field networks help the network to more effectively factor out articulation from non-rigid components. 
We further extend our model to multiple shape setting and show comparable performance to concurrent work LEAP~\cite{LEAP:CVPR:21}.

We believe that Neural Generalized Implicit Functions open several interesting research directions.
Since currently we have a clothing-specific model, it will be useful to extend this approach such that it can animate multiple clothing using the same model. 
It would be also valuable to learn temporal correspondences implicitly during learning. Finally, since Neural-GIF produces signed distance fields, we want to leverage them for fast collision and contact computation in human-object and scene interactions.

{\small\paragraph{{\bf Acknowledgements:}}This work is funded by the Deutsche Forschungsgemeinschaft (DFG, German Research Foundation) - 409792180 (Emmy Noether Programme,
project: Real Virtual Humans) and a Facebook research award. }

\clearpage
{\small
\bibliographystyle{ieee_fullname}
\bibliography{egbib}
}

\end{document}